\documentclass{llncs}
\usepackage{makeidx}  
\usepackage[colorlinks]{hyperref}
\usepackage{siunitx}
\usepackage{tabularx}
\usepackage{graphicx}
\usepackage{subcaption}
\usepackage{enumitem}

\usepackage{etoolbox} 
\makeatletter
\patchcmd{\ps@headings}{\rlap{\thepage}}{}{}{}
\patchcmd{\ps@headings}{\llap{\thepage}}{}{}{}
\makeatother
\begin{document}
%
%

%
\mainmatter              
\title{A Social Robot with Inner Speech for Dietary Guidance}
\titlerunning{Hamiltonian Mechanics}  
%
\author{
Valerio Belcamino\inst{2} \and Alessandro Carfì\inst{1} \and Valeria Seidita\inst{2}\and Fulvio Mastrogiovanni\inst{1} \and Antonio Chella\inst{2}
}
\authorrunning{Ivar Ekeland et al.} 
%
\tocauthor{Valerio Belcamino, Alessandro Carfì, Valeria Seidita, Fulvio Mastrogiovanni and Antonio Chella}

\institute{TheEngineRoom, Department of Informatics Bioengineering, Robotics and System Engineering, University of Genoa, Genoa, Italy,\\
\email{valeriobelcamino@gmail.com},
\texttt{}
\and
Department of Engineering, University of Palermo, Palermo, Italy}

\maketitle              

\begin{abstract}
We explore the use of inner speech as a mechanism to enhance transparency and trust in social robots for dietary advice. In humans, inner speech structures thought processes and decision-making; in robotics, it improves explainability by making reasoning explicit. This is crucial in healthcare scenarios, where trust in robotic assistants depends on both accurate recommendations and human-like dialogue, which make interactions more natural and engaging.
Building on this, we developed a social robot that provides dietary advice, and we provided the architecture with inner speech capabilities to validate user input, refine reasoning, and generate clear justifications. The system integrates large language models for natural language understanding and a knowledge graph for structured dietary information. By making decisions more transparent, our approach strengthens trust and improves human–robot interaction in healthcare. We validated this by measuring the computational efficiency of our architecture and conducting a small user study, which assessed the reliability of inner speech in explaining the robot's behavior.
\keywords{Social Robotics, Cognitive Architectures, Human-Robot Interaction (HRI), Large Language Models (LLM)}
\end{abstract}
\section{Introduction}\label{sec:intro}
Human-robot interaction (HRI) has evolved from basic task execution to more sophisticated systems, including robotic companions and chatbots that can engage in natural, human-like conversations~\cite{nl}. As robots increasingly serve as partners and assistants across various domains, their ability to communicate effectively~\cite{communication}, reason transparently~\cite{transparency}, and inspire trust~\cite{trust} becomes an essential factor. This is particularly true in high-stakes applications such as healthcare~\cite{trust_health}, where the consequences of miscommunication or opaque decision-making can have significant repercussions on patient well-being. In such settings, ensuring that robotic systems are not only functionally competent but also cognitively relatable~\cite{relatable,relatable2} and trustworthy is a critical design goal.
Healthcare scenarios present unique challenges and opportunities for HRI~\cite{healthcare}. Patients and caregivers alike require systems that can articulate their reasoning, adapt to complex personal needs, and operate with a high level of transparency to promote trust ~\cite{healthcare_personality}. In environments where adherence to treatment plans or dietary regimes is crucial, the ability of a robot to explain its actions in clear, human-like terms can substantially influence user engagement and compliance. Here, trust is not merely a function of reliable performance; it is deeply intertwined with robot communication and cognitive capabilities~\cite{trust_health2}.
Trust in robot companions is often built when users can attribute human-like features to these systems, both in terms of physical embodiment~\cite{embodiment} and behavioral expression~\cite{behavior}. Empirical studies in HRI have consistently shown that both verbal communication~\cite{verbal_trust} and non-verbal cues~\cite{gaze} like gestures and gaze play crucial roles in establishing trust~\cite{nonverbal_trust}. 
Moreover, trust is closely related to the psychological concept of the theory of mind: users are more inclined to trust a robot when they can discern clear patterns in its operations and understand its decision-making processes~\cite{transparency2}. Providing transparent explanations at both the operational and reasoning levels not only facilitates accurate mental models of the robot but also reinforces user confidence. In human cognition, this intricate interplay between communication and reasoning can be represented by inner speech~\cite{is2}.

At the heart of human cognition lies the phenomenon of inner speech, a covert dialogue that guides self-regulation, decision making, and working memory~\cite{innerspeech}. Vygotsky’s theory~\cite{vygotsky} posits that inner speech arises from the repeated externalization and subsequent internalization of caregiver instructions, evolving into an internal monologue that aids in planning and problem-solving. Complementary working memory models~\cite{is_working_memory} suggest that inner speech functions as a cognitive mechanism for holding and manipulating task-relevant information, much like keeping a grocery list in mind while shopping. This dual role, both as a self-regulatory tool and a means of sustaining temporary information, underscores the critical function of inner speech in human thought processes~\cite{al2006sociocultural}.
Translating the concept of inner speech to robotics represents an emerging frontier with significant potential. Although the literature remains relatively sparse, recent breakthroughs in large language models have drastically improved natural language generation~\cite{llm}, allowing machines to produce dialogue that closely mimics human communication. This advancement not only narrows the gap between human-machine and human-human interactions but also opens new avenues to improve trust through improved language capabilities~\cite{llm_trust}. In our previous work~\cite{table}, we demonstrated that exposing users to a robot’s inner speech, in the form of overt self-talk, can enhance perceptions of animacy and intelligence, key attributes that underpin trust.

Building on these insights, we propose a novel cognitive architecture that integrates an inner speech module into the robotic system. This module not only orchestrates software components for coherent and context-sensitive decision making but also generates internal dialogue that, when externalized, provides users with transparent insights into robot reasoning. 
In our architecture, we leverage large language models (LLMs) for both inner and outer speech, enabling the robot to communicate in a natural, human-like manner. However, while LLMs are powerful in generating fluent, context-aware responses, they are fundamentally ``stochastic parrots"~\cite{stochastic_parrot}—producing plausible text without true understanding. To address this limitation, we balance the flexibility of LLM-driven stochastic decision making with the deterministic reasoning of symbolic methods, such as ontology-based knowledge graphs and logic programming.
As an initial validation for our approach, we conducted experiments including a detailed latency analysis of the system’s core modules and a user study assessing the efficacy of inner speech in conveying both the user’s request and the robot’s intended course of action.

\section{Methodology}
\label{sec:methodology}
\begin{figure}[t!]
\centering
\includegraphics[width=0.65\textwidth]{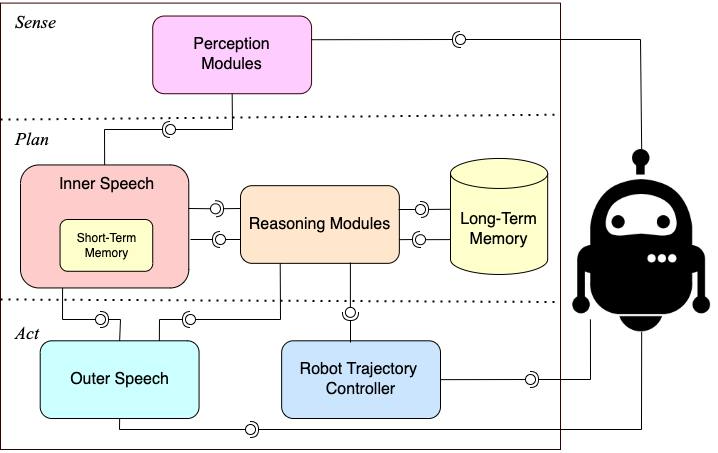}
\caption{Overview of the proposed cognitive architecture. The Perception Modules process sensory input and provide information to the Inner Speech Module, which enables self-reflection and contextual reasoning while maintaining a Short-Term Memory for temporary information retention. The Reasoning Modules integrate symbolic and stochastic reasoning to support decision-making. A separate Long-Term Memory stores persistent knowledge, which the reasoning process can access. Finally, the Outer Speech Module externalizes selected reasoning steps to enhance transparency, while the Robot Trajectory Controller executes physical actions.}
\label{fig:architecture_general}
\end{figure}

The proposed architecture is shown in Figure~\ref{fig:architecture_general} and follows the Sense-Plan-Act (SPA) paradigm, a widely adopted kind of architecture in robotics that structures operation into three key phases: perception (sense), reasoning and planning (plan), and execution (act). 
The \textit{sense} layer is handled by the perception modules, which process sensory data from the environment. These modules extract relevant features, such as object recognition, spatial awareness, or user interactions, depending on the specific robotic application. The structured information generated at this stage serves as the foundation for subsequent reasoning and decision-making processes in the Plan phase.

The \textit{plan} phase consists of two core components: the inner speech (IS) module and the reasoning modules. The IS module, built on a large language model, orchestrates the flow of information among various components to produce coherent and well-founded decisions. Unlike conventional approaches, our inner speech module generates a natural language self-dialogue in a note-taking style, providing a detailed account of each reasoning step. This process decomposes complex decisions into intermediate sub-tasks, in a manner similar to chain-of-thought~\cite{cot} and tree-of-thought~\cite{tot} approaches, iteratively refining the output to improve internal consistency and extending the context window of the model.
A key feature of the inner speech module is its dynamic short-term memory system, which retains fragments of prior inner speech, intermediate reasoning steps, and contextual data from other modules. This is achieved by concatenating these fragments into the prompt, allowing the module to reference and build upon previous information during reasoning, allowing the system to incorporate past information into its decision-making process, ensuring that decisions are continuously shaped by accumulated experience and external inputs. 
Moreover, the natural language explanations produced by the inner speech module play a key role in promoting user transparency. When this self-dialogue is disclosed, it offers a clear and accessible account of the robotic advisor’s decision-making policy, thereby explicating the rationale behind each recommendation~\cite{llm_healthcare}. By enabling users to understand the underlying reasoning, the system not only enhances the perceived reliability of its recommendations but also facilitates a more informed and confident user interaction.
Overall, the dual functionality of the inner speech module—improving internal reasoning through structured self-dialogue and enhancing external explainability via transparent communication—constitutes a fundamental mechanism in our dietary advisory framework, ultimately contributing to both improved system performance and increased user trust.
Alongside inner speech, the reasoning modules encompass both symbolic inference mechanisms based on logical programming and stochastic models such as LLMs. The combination of these paradigms allows for:
symbolic reasoning, which enforces rule-based logical constraints and verifiable decision-making,
Probabilistic reasoning, which enables flexible adaptation to novel scenarios where predefined rules may be insufficient.
This hybrid approach ensures that decisions are not only contextually adaptive but also grounded in structured formal knowledge where necessary. Additionally, the reasoning modules can access a long-term memory, which serves as a persistent knowledge base containing world models, learned experiences, and stored facts. Unlike short-term memory, which dynamically updates within an active session, long-term memory retains information across multiple interactions and can be modified by symbolic reasoning processes as needed.

Finally, there is the \textit{act} layer, where decisions generated during planning are transformed into concrete actions. This layer includes the Outer Speech Module and the Robot Trajectory Controller.
The outer speech module verbalizes selected portions of the inner speech process, allowing users to gain insight into how decisions were reached. While not all internal reasoning is externalized, strategically disclosing key aspects enhances transparency and enables users to detect potential errors preemptively. The generated speech follows a structured format resembling human note-taking during problem-solving, presenting sequences of actions and observations in a logical flow.
The robot trajectory controller translates high-level decisions into physical actions by generating movement trajectories based on planned objectives. This module ensures that motor commands align with both reasoning outputs and environmental constraints, allowing the robot to perform tasks safely and efficiently.

\section{Implementation}
\label{Implementation}

\begin{figure}[t!]
\centering
\includegraphics[width=0.65\textwidth]{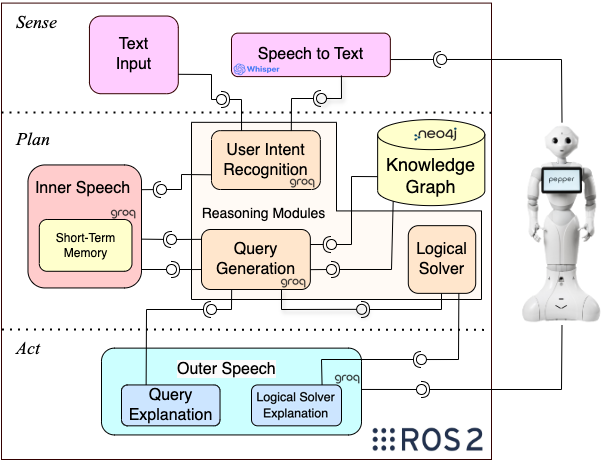}
\caption{Overview of the proposed implementation. The Sense layer captures user input via text or a speech-to-text module, while the Plan layer coordinates dialogue management (Inner Speech), intent recognition, query generation, and logical reasoning through the knowledge graph and logical solver. Finally, the Act layer delivers responses to the user through Outer Speech and explanatory modules, closing the interaction loop.}
\label{fig:architecture_impl}
\end{figure}

Building on the general architecture introduced in Section \ref{sec:methodology}, we now detail our prototype implementation, which was designed to support a robotic advisor that assists users in following a dietary plan. Every choice the system makes is supported by a thorough explanation, which promotes user trust given the delicate nature of healthcare.
Figure~\ref{fig:architecture_impl} shows our implementation, which was developed on ROS2 Humble using Python and deployed on Ubuntu 22.04. The physical embodiment is provided by the Pepper robot, selected for its inherent interactive capabilities. At the core of our system is a knowledge graph (KG) implemented in Neo4j, which models three main entities: users, dishes, and allergens. The KG also defines several relations among these entities to support personalized dietary recommendations. Specifically, the relation "\textit{has allergen}" links a dish to the allergens it contains, while "\textit{is allergic to}" associates users with the allergens they must avoid. Additionally, the relation "\textit{has nutritional needs}" connects each user to a set of dietary requirements, specifying target values for calories, carbohydrates, proteins, and fats. This structure allows the system to effectively assess the suitability of dishes for individual users, ensuring both allergen safety and nutritional alignment. The KG enables three primary operations. First, \textit{User Insertion} allows the addition of new users along with their nutritional profiles and allergies. Second, \textit{Dish Information} provides detailed nutritional data and, when a specific user is referenced, includes allergen warnings. Finally, \textit{Meal Preparation} generates optimal dish combinations tailored to fit the user's nutritional needs.

Building on this foundation, our architecture consists of three key layers: perception, planning, and execution, each responsible for different aspects of interaction and decision-making. At the perception level, our system employs a speech-to-text module based on a locally deployed, small version of OpenAI’s Whisper\footnote{\scriptsize{\href{https://openai.com/index/whisper/}{openai.com/index/whisper}}}, which captures and processes verbal commands. To ensure versatility, we also provide a text-based interaction channel that serves users who prefer or require a non-verbal interface. In the planning layer, user prompts are initially handled by the intent recognition (IR) module. This module leverages a LLM to classify requests among the three primary operations: user insertion, dish information, and meal preparation. A railguard mechanism within the IR module filters out queries that are either off-topic or unsupported by the current implementation. In parallel, the IR module extracts all parameters pertinent to the requested action (e.g., dish names, user identifiers, or nutritional preferences) and forwards them to the inner speech (IS) module.
The IS module supervises the extracted parameters by verifying whether the user’s input contains sufficient information to execute the desired operation. In cases where key details are missing (e.g., a request for user insertion that only includes a name), the IS module initiates a replanning process. This involves activating the Outer Speech module, which prompts the user for additional data via the perception module. Conversely, if the prompt is complete, the IS module triggers the Query Generation module, which translates the refined request and its associated parameters into one or more precise Cypher queries.
Once generated, these queries are executed on the Neo4j database. If a query fails or returns incomplete data, the IS module reactivates the replanning process, asking the user to clarify or refine their request. Otherwise, the query results, along with the original query, are passed to the Query Explanation module, which produces a natural language summary of how the filtering or retrieval was performed. In the case of the Meal Preparation operation, the system must not only filter dishes to exclude harmful allergens but also identify an optimal combination of dishes that meets the user’s nutritional goals. To this end, the filtered dish list and the user’s nutritional requirements are fed into a logical solver implemented with Prolog and Clingo. By employing symbolic reasoning, the solver searches for dish combinations that minimize the difference between the user’s target nutritional values and the sum of the selected dishes, allowing a ±10\% threshold to remove combinations that are too distant from the target. 
To enhance transparency, the generated meal plans are processed by the Solver Explanation module, which presents each solution to the user. This module provides a structured breakdown of the recommended dishes, their calorie and macronutrient content, and the deviation from the user’s nutritional targets. By clearly communicating the reasoning behind each recommendation, the system ensures that users can make informed decisions.

The LLM-based modules in our system, intent recognition, inner speech, query generation, and both outer speech modules, are implemented using Groq’s free-tier APIs, utilizing the LLaMA 3 (70B) model. These modules enable natural language understanding, reasoning, query formulation, and response generation within the robotic assistant.
To ensure appropriate behavior, temperature values were empirically selected based on each module’s function. Intent recognition and query generation require strict adherence to syntax and predictable outputs, so the temperature is set to 0.0 to maximize determinism. In contrast, the outer speech modules, responsible for generating natural language responses for user interaction, benefit from slight linguistic variability, leading to a temperature setting of 0.25. Inner Speech, which produces structured reasoning in a note-taking format, represents an intermediate case. To balance coherence with natural expression, its temperature is set to 0.1. These values were chosen as a trade-off between adherence to the prompt and linguistic variability, ensuring both precision in structured outputs and naturalness in dialogue-based interactions.
Since no fine-tuning was performed, all LLM-based modules rely on few-shot prompting, allowing the model to generate more contextually relevant responses by incorporating examples into the input prompt. This approach leverages the model’s expanded context window to improve accuracy in decision-making and user interactions. The complete implementation, including code and example prompts, is publicly available in our repository\footnote{\scriptsize{\href{https://github.com/ValerioBelcamino/unipa_inner_speech}{https://github.com/ValerioBelcamino/unipa\_inner\_speech}}}.

\section{Results}
\label{Results}

\begin{figure}[t!]
\centering
\includegraphics[width=\textwidth]{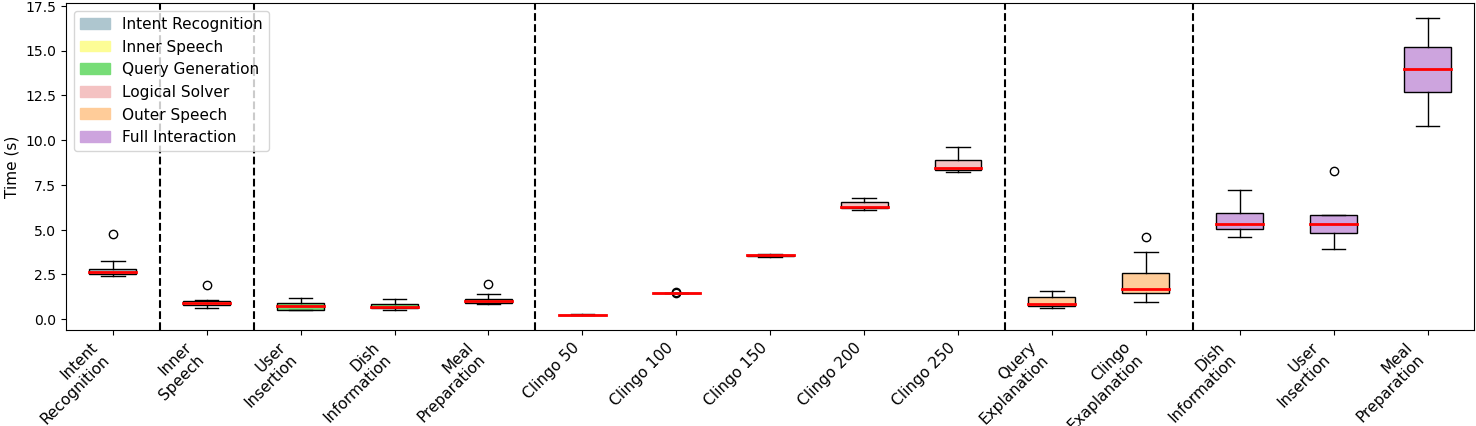}
\caption{Time analysis for different system components and operations. The boxplot illustrates the time required (in seconds) for key processes, including intent recognition, inner speech, query generation, Clingo-based reasoning with varying number of items, and explainability mechanisms. The final three categories represent the full interaction time for dish information retrieval, user insertion, and meal preparation.}
\label{fig:time}
\end{figure}

A key requirement for deploying social robots in healthcare applications is ensuring that system responses are both transparent and efficient. While explainability mechanisms, such as inner speech and symbolic reasoning, enhance user trust, they should not introduce excessive delays that could hinder real-time interaction. Large language models (LLMs) provide flexibility and human-like language manipulation, but they are generally slower. Thus, we conducted a timing analysis on the core components of the system, measuring their execution times across different operations. The results, presented in Figure~\ref{fig:time}, provide insights into the computational costs associated with each module separately and with a the full interaction for the three supported tasks. Each test was performed with 20 different input prompts.

The execution times for LLM-based modules (i.e., intent recognition, inner speech processing, query generation, and explainability) depend primarily on the number of input and output tokens. Our analysis shows that intent recognition exhibits an average latency of 2.58\si{\second} (SD = 0.30), reflecting the need to classify user intent while handling diverse input phrasing. This higher processing time may be attributed to the fact that the prompt for intent recognition is longer, richer in detail, and includes multiple examples to ensure accurate classification. In contrast, inner speech, which generates structured reasoning in a note-taking style, operates significantly faster, with an average execution time of 1.02\si{\second} (SD = 0.20). 
The translation from natural language to Cypher queries is highly efficient across all operations. Query generation for user insertion completes in 0.74\si{\second} (SD = 0.05), while dish information queries are slightly faster at 0.73\si{\second} (SD = 0.03), reflecting the straightforward mapping of user input to database retrieval commands.
Meal preparation queries, however, require more processing time, averaging 1.28\si{\second} (SD = 0.11). This increase is due to the fact that, unlike the other operations, meal preparation involves generating two separate queries: one to retrieve the user’s nutritional requirements and another to filter the available dishes based on the user’s allergies. 
Explainability modules introduce additional but controlled overhead, depending on whether the explanation pertains to database queries or symbolic reasoning outputs. The explanation of the queries can be executed in 1.00\si{\second} (SD = 0.12), while the explanations for the logical reasoner require more time at 2.13\si{\second} (SD = 1.20). Again, the time difference between this two modules can be attributed to the different number of tokens; in fact, in the case of the Clingo solver, it may generate multiple answers, thus extending the number of input tokens and the resulting natural language explanation. Similarly, the standard deviation is higher since the number of solution is variable.

In contrast to LLM-based modules, the symbolic reasoning based on Clingo exhibits a more pronounced dependency on problem complexity. Since meal preparation involves finding an optimal combination of dishes that aligns with a user's nutritional needs while avoiding allergens, the computational cost scales with the number of available dishes. When reasoning over a set of 50 dishes, execution time remains low at 0.25\si{\second} (SD = 0.01), but as the number increases to 100, 150, and 200 dishes, processing times grow to 1.46\si{\second} (SD = 0.01), 3.54\si{\second} (SD = 0.02), and 6.37\si{\second} (SD = 0.05), respectively. The most complex scenario, which involves 250 dishes, requires 8.63\si{\second} (SD = 0.22), reflecting the increased search space required to generate balanced meal plans. It is worth noting that the optimizer performs pruning on the problem by grouping items with similar values and discarding solutions that deviate excessively from the target criteria. These simplifications slightly reduce the complexity of the problem, which would otherwise grow exponentially with the number of items. Additionally, these results have been obtained on a laptop with an Intel i5 7th gen processor and 8GB of ram.
To provide a fair comparison between single-module execution times and full interaction performance, the values reported in Figure~\ref{fig:time} correspond to scenarios where the user prompt was already complete, avoiding additional replanning steps. Under these conditions, both user insertion and dish information retrieval maintain relatively stable execution times, averaging 5.56\si{\second} (SD = 2.74) and 5.58\si{\second} (SD = 0.88), respectively. For meal preparation, performed via Clingo-based optimization over a dish database of 150 items, the execution time exhibits greater variability, with an average of 13.89\si{\second} (SD = 6.36) and upper bounds extending to approximately 17\si{\second} in more complex cases.
However, when the input lacks necessary details, the intent recognition and inner speech modules can be invoked multiple times to request clarification, leading to extended interaction times. In scenarios where two replanning iterations are required, dish information retrieval and user insertion can reach up to 15\si{\second}, while meal preparation can exceed 20\si{\second} due to the additional processing and reasoning steps needed to refine the request. This highlights the impact of incomplete prompts on overall system responsiveness and underscores the role of inner speech in guiding users toward structured interactions.

The results indicate that the overall interaction time remains satisfactory, as the introduction of the inner speech module did not create a bottleneck for the system. In cases requiring replanning, the higher latency is justified since it includes user response time. Furthermore, the modular interplay grants flexibility, enabling the system to adapt to diverse requests and be fine-tuned by modifying prompts and examples.
However, interaction times remain relatively high. While sufficient for simple, one-shot user queries~\cite{llm_alexa}, they fall short of enabling fluid, natural dialogue between humans and robots. As an example, if the inner speech module is disclosed to the user, the system's response time is around 3.60s, as it only involves intent recognition and inner speech inference. While this duration is closer to what users typically experience with home assistants, it is still far from the responsiveness of natural human-human dialogue.
Although Groq APIs are among the fastest available for LLM inference~\cite{groq}, network delays still contribute to overall latency. A potential solution is deploying a locally hosted model, which could eliminate network overhead and reduce inference time to around 100ms~\cite{local}. Another bottleneck stems from the few-shot prompting technique. To ensure a structured output—such as in intent recognition—the input prompt requires an extensive problem description and multiple examples covering different cases. Modules that rely on strict output formatting would benefit significantly from fine-tuning, as they do not require the full natural language capabilities of the base model.
Finally, in social Human-Robot Interaction (HRI) literature, response delays are often mitigated through human-like behaviors, including gestures, conversational fillers, and backchanneling~\cite{backchanneling2}. These strategies could help enhance perceived responsiveness and improve the naturalness of the interaction.

\begin{figure}[t!]
\centering
\includegraphics[width=\textwidth]{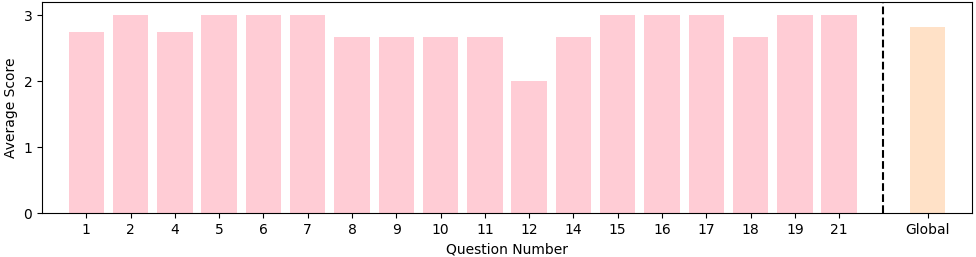}
\caption{The plot shows the average scores for each question of the questionnaires, reflecting participants' understanding of i) the robot's intent, ii) the required parameters, and iii) the completeness of the initial prompt. Questions 3, 13, and 20 were excluded from the analysis because metrics ii and iii cannot be applied to out-of-scope samples. The final column represents the overall average score.}

\label{fig:questionnaires}
\end{figure}

To assess the ability of the inner speech module to convey the intended procedure of the robot, we conducted an experiment with 14 participants. Each participant completed three questionnaires, each containing 7 inner speech samples, for a total of 21 unique examples. The 7 samples in each questionnaire were selected to ensure an even distribution across the four supported interaction types, including out-of-scope prompts. In some cases, the fragments consisted of multiple segments of robot reasoning resulting from incomplete user inputs. Participants were asked to describe, in natural language, both the user's request and the robot's intended course of action solely based on the inner speech prompt, without receiving the original input or the generated outer speech. The participants' understanding was evaluated using a questionnaire based on the following three metrics:

\begin{enumerate}[label=\roman*)] \item Understanding the robot's intent: whether the participant correctly identified what the robot was trying to accomplish. \item Understanding the required parameters: whether the participant correctly identified the key parameters needed for the robot to complete its task. \item Evaluation of the completeness of the initial prompt: whether the participant correctly assessed whether the user's initial input contained all the necessary information for the robot to act. \end{enumerate}

Each response was assigned a score of one point for each of the three metrics. In Figure~\ref{fig:questionnaires}, we present the average score across participants for each question, shown in the corresponding columns. The overall evaluation yielded a global average score of 2.81, which is a satisfactory result. As a general observation, the robot’s intent is almost always correctly captured by the participants, as is the identification of missing parameters in incomplete prompts. The global value reflects the uniformity of the responses across most questions, with the notable exception of question 12, which only reached a mean score of 2.00. This particular question involved a misuse of the system, where a user first asked an incomplete question regarding the insertion task and then, when prompted for additional information, changed topic to request details about a specific dish. We excluded questions 3, 13, and 20 from the analysis, as we considered the use cases of our application to be well defined, not requiring additional explanations for out-of-scope prompts. Consequently, users lacked sufficient information to evaluate metrics (ii) and (iii). Nevertheless, they were still able to correctly identify the out-of-scope prompts (i.e., metric (i)). While this is a positive result, the approach may pose a limitation for detecting false negatives—user prompts that are incorrectly classified as out of scope.
Since this work is part of the national Italian project ADVISOR, all trials were conducted in Italian with Italian speakers. A supervisor manually reviewed all responses to confirm that they accurately reflected the intended initial prompt. The consistent accuracy of the participants’ interpretations highlights the effectiveness of the inner speech module in transparently communicating the robot’s reasoning and planned actions.
These findings suggest that internal self-dialogue can enhance user understanding in robotic decision-making, a key aspect for applications in sensitive domains such as healthcare. The questionnaires, along with the answers and the English translation, are available in our repository\footnote{\scriptsize{\href{https://github.com/ValerioBelcamino/inner_speech_questionnaries}{https://github.com/ValerioBelcamino/inner\_speech\_questionnaries}}}.
\section{Conclusions}
We proposed an architecture for a social robot assistant that leverages inner speech to plan its actions while providing contextual reasoning in natural language. When disclosed to the user, this internal dialogue improves transparency by making the robot’s decision-making process more understandable, which can be particularly useful in healhcare scenarios. We implemented our architecture in a system designed to help users follow dietary plans, integrating LLM-based modules with logical programming and a knowledge graph.

We evaluated the latency of the system, showing that, while response times are still not comparable to human dialogue, they remain satisfactory for one-shot interactions. In addition, a supplementary experiment demonstrated that participants could accurately interpret both the user's request and the robot's intended course of action solely based on inner speech prompts, further validating the module’s role in enhancing transparency. Future work will explore reducing latency by deploying locally hosted models and fine-tuning structured output modules. Additionally, we plan to investigate methods commonly used in social HRI, such as gestures, conversational fillers, and backchanneling, to enhance perceived responsiveness and human-likeness. Finally, a user study may be extremely useful to analyze the impact of inner speech capabilities on trust and engagement in human-robot interaction.

\section*{Acknowledgement}
This research is partially supported by the Italian government under the National Recovery and Resilience Plan (NRRP), Mission 4, Component 2, Investment 1.5, funded by the European Union NextGenerationEU programme, and awarded by the Italian Ministry of University and Research, project RAISE, grant agreement no. J33C22001220001. In addition, the ADVISOR project was founded by European Union - NextGenerationEU, Mission 4 Component 1 CUP E53D23016260001.



\bibliographystyle{IEEEtran}

\bibliography{bibliography}

\end{document}